# Emotive Response to a Hybrid-Face Robot and Translation to Consumer Social Robots

Maitreyee Wairagkar, Maria R. Lima, Daniel Bazo, Richard Craig, Hugo Weissbart, Appolinaire C. Etoundi, Tobias Reichenbach, Prashant Iyenger, Sneh Vaswani, Christopher James, Payam Barnaghi, Chris Melhuish, and Ravi Vaidyanathan

*Abstract*— We introduce the conceptual formulation, design, fabrication, control and commercial translation with IoT connection of a hybrid-face social robot and validation of human emotional response to its affective interactions. The hybrid-face robot integrates a 3D printed faceplate and a digital display to simplify conveyance of complex facial movements while providing the impression of three-dimensional depth for natural interaction. We map the space of potential emotions of the robot to specific facial feature parameters and characterise the recognisability of the humanoid hybrid-face robot's archetypal facial expressions. We introduce pupil dilation as an additional degree of freedom for conveyance of emotive states. Human interaction experiments demonstrate the ability to effectively convey emotion from the hybrid-robot face to human observers by mapping their neurophysiological electroencephalography (EEG) response to perceived emotional information and through interviews. Results show main hybrid-face robotic expressions can be discriminated with recognition rates above 80% and invoke human emotive response similar to that of actual human faces as measured by the face-specific N170 event-related potentials in EEG. The hybrid-face robot concept has been modified, implemented, and released in the commercial IoT robotic platform Miko (My Companion), an affective robot with facial and conversational features currently in use for human-robot interaction in children by Emotix Inc. We demonstrate that human EEG responses to Miko emotions are comparative to neurophysiological responses for actual human facial recognition. Finally, interviews show above 90% expression recognition rates in our commercial robot. We conclude that simplified hybrid-face abstraction conveys emotions effectively and enhances human-robot interaction.

## I. INTRODUCTION

### A. Research context

Affective social robots are gaining increasing interest in research and social applications. However, achieving smooth human-robot interaction still has significant challenges such as robots to become trustworthy to humans through the incorporation of emotional compatibility in their interactions [1-4]. Humanoid social robots provide means to investigate social cognition, engage with, and support human mental health.

It is now well-accepted that humans respond better to robots that behave empathetically towards them, which involves the capacity to recognise emotion and respond accordingly [5-11]. Pioneering work, in particular by Brazeal and Ishiguro [4, 12-14], grounded this field of study with a very strong body of literature now available on affective human-robot communication (see [7, 8, 15, 16] for recent surveys). Industry translation has also begun in areas such as service and hospitality, highlighted by the opening of the Henn-na Hotel in Nagasaki in 2015, though challenges in reliability and acceptance by humans remain unresolved [17].

For a successful natural human-robot collaboration, social robots must adopt a multimodal approach with capability to show facial expressions, speech, gestures, access online knowledge, understand context and intent, be aware of surroundings and adapt their behaviour accordingly. This can be achieved by connecting social robots to the Internet of Things (IoT) and cloud services to enhance their social and emotional capabilities [18-20] while improving security [21]. IoT based social robots have been proposed for use in education [22], special needs education [23], in healthcare for cognitive therapy [24] and assistive and care services [19, 21, 25] all of which require an emotional connection, empathy and trust with users. There has been relatively less research on embedding sensory information from IoT to develop interactive social robots for detecting and responding to emotions using visual expressions.

The expression of realistic robotic emotion calls for the fusion of numerous disciplines, including systems design, user experience, artificial intelligence and electro-mechanical hardware integration. Ekman and Friesen [26] described and tested the assumption of universality in human interaction, asserting six primary emotions: happy, sad angry, afraid, surprise and disgust. Ekman proposed that all human expressions can be represented by a combination of six basic expressions according to the Facial Action Coding System (FACS) [27]. Thus, emotions can be characterised by using continuous scales or a three-dimensional 'affect space' to

Support of the UK Dementia Research Institute Care Research and Technology Centre (DRI-CRT)*, UK EPSRC (EP/F01869X/1)* UK GCRF* European Commission (CHRIS FP7-215805) ^ and Emotix Inc & (*PI: R Vaidyanathan; ^ PI: C. Melhuish; & Co-PIs: S Vanaswami, P Iyenger).

M. Wairagkar, M R. Lima, and R Vaidyanathan are with Dept of Mechanical Engineering, Imperial College London and the UK Dementia Research CRT, London, UK (r.vaidyanathan@ic.ac.uk)

P Barnaghi is with Dept of Brain Sciences, Imperial College London and the UK Dementia Research Institute CRT, London, UK

H Weissbart, T Reichenbach are with Dept of Bioengineering, Imperial College London, London, UK

D Bazo is with the Dept of Media Arts, University of California, Santa Barbara, USA

A Etoundi, R Craig, and C Melhuish are with the Bristol Robotics Laboratory, University of West England/University of Bristol, UK.

C James is with the Dept of Bioengineering, University of Warwick, UK

P Iyengar and S Vaswani are with Emotix Inc, Mumbai, India

mathematically represent facial expressions using axes, such as arousal, valence, and stance. The configuration of a face is also important for its recognition. When designing humanoid social robots, care should be taken to avoid reaching the 'uncanny valley' [28], a state in which robot's appearance becomes eerily human-like leading to negative emotional response including revulsion and discomfort by humans.

Expressive behaviour of robotic faces, however, also require complex mechatronic design and control [29]. Full facial actuation and control, despite many successful applications, remains an issue limiting widespread adoption. This has motivated a significant body of research aimed at simplification for cost and ease of use (e.g. [30-33]). A plateau is also being approached regarding the degrees of freedom that can be conveyed by robot's facial expressions; additional human-like emotive conveyance such as pupil dilation, which is often overlooked, can reveal the emotional state and intention behind expression present further challenges to be addressed.

Finally, while human-robot affective communication is a maturing field, empirical assessment of human-robot affective interactions has not been fully addressed. There is a paucity of research in human emotional response to the robot's facial expressions as compared to affective communication with virtual avatars [15]. Recent investigations, such as the implementation of the FACS [32] offer significant potential for empirical assessment, however direct analysis of physiological response also holds significant potential in this realm.

It is interesting to investigate the effects of robot interactions on human physiological responses using electroencephalography (EEG) or functional magnetic resonance imagining (fMRI) to assess engagement. Wang's [33] pioneering work compared human-robot interaction to human-human interaction using detailed fMRI and showed that human-robot interaction elicited stronger feelings of eeriness than human-human interaction, though their experiments were limited to colour images. Recognising emotion and facial expressions involve several task-specific neuron sources [34] and different parts of the brain activate in response to different facial expressions [35]. The right hemisphere of the brain plays an important role in processing emotions [36, 37]. A common method of assessing neurophysiological response in EEG is by studying Event-Related Potentials (ERP) which are positive or negative voltage deflections time-locked to stimulus onset [38]. N170, a negative potential occurring 170 ms after stimulus onset is a well-documented face-specific ERP component that occurs in response to visual face stimulus [39, 40]. The N170 ERP component has been attributed to the face-specific structural encoding stage occurring before the recognition of face and is reported to be unaffected by emotional content within expressions[37, 41]. Different methods such as [42] can be used to isolate regions of brain activity in response to ERPs. These established neurophysiological studies also hold intriguing potential for empirical quantification of human response to a robot's expressions [43].

*B. Scope of work*

The goal of our ongoing work is to develop a practical approach to design simplified affective robots capable of emotive conveyance and to validate this approach through quantification of robotic interaction with humans using conscious (behaviour) and subconscious (neurophysiological) response measures. This work is motivated by pragmatic considerations for widespread use where highly actuated faces are not viable due to cost and maintenance constraints. We aim to design simpler robotic faces that evoke comparable human-like interaction. Facial expressions, in particular, are typically conveyed through highly advanced digital graphical platforms or complex physical actuation displaying realistic facial movement which are not only very complex to program and maintain but are also sometimes off-putting to users. Identification and canonical abstraction of features of a robotic face that evoke similar responses from humans as a fully actuated (ideally an actual human) face have significant potential to enable broader and more accessible use of social robots [7].

In this study, we extend our past research [43, 44] to introduce a complete hybrid-face affective robotic system to convey human-like facial emotions without the complexity of full facial actuation. We present the robot design, modelling of affect space emotions, validation through empirical (neurophysiological) assessment of human response during robot interaction, and simplification of the concept to deliver a complete commercial IoT robot system ('Miko' – 'my companion') that is in use today for affective robot interaction with children. This development provides a basis for the larger goal of developing mechanically simple platforms for human-robot engagement as well as a method to quantify, physiologically, human response to affective robots.

Our aim is to 1) demonstrate the capacity of our two robot systems with simplified digital affective face to visually convey emotions to a human observer and 2) quantify their conscious and subconscious responses to these emotions while demonstrating similarity between neurophysiological response to human and robotic emotive conveyance, ultimately to answer how effective are these robots in conveying emotions?

The scope of this study encompasses: 1) the development of a hybrid-face humanoid robot capable of emotive response by modelling of emotion affect space, and integrating pupil dilation in expressions; 2) testing of hybrid-face through mapping human response to the robot qualitatively (interview/feedback) and quantitatively (EEG); 3) applying the findings from the hybrid-face robot to develop the commercial affective IoT social robot Miko, and 4) testing emotion conveyance of Miko in the same manner established by hybrid-face robot qualitatively (interview/feedback) and quantitatively (EEG). In this paper, we present the successful development of the simplified hybrid-face affective robot, its translation to a commercial robot Miko and validate their affective face designs through qualitative analysis of human physiological response.

II. HYBRID-FACE AFFECTIVE ROBOT

*A. Design of Hybrid-Face Affective Robot*

The hybrid-face robot shown in Fig. 1(a) combines a digital face with a static 3D printed human visage-like structure (Fig. 1(b)). It is designed to provide the flexibility of a digital countenance with some of the benefits of a full-featured, fully actuated face.

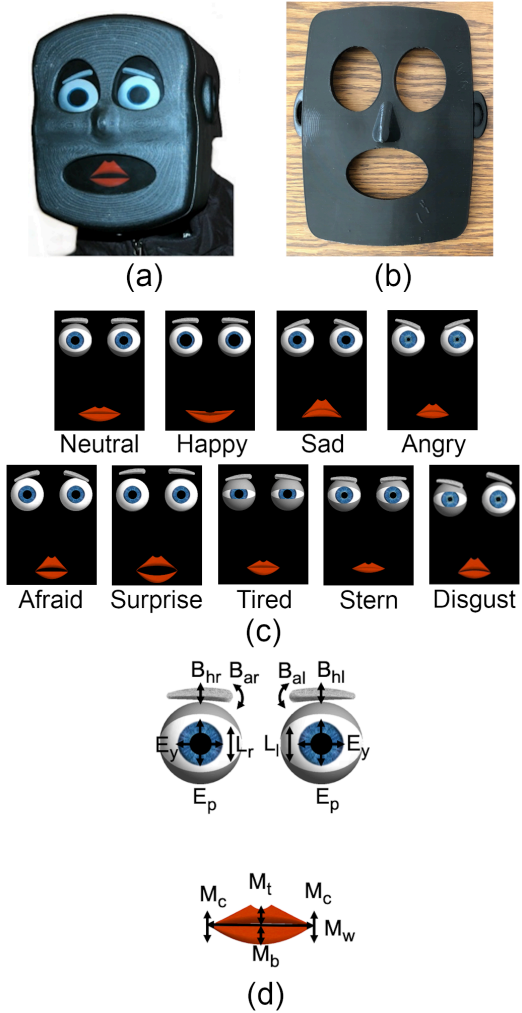

Fig. 1. Hybrid-facer robot and its facial expressions. (a) The hybrid-face robot with a faceplate and digital display (b) 3D printed faceplate. (c) Facial expressions of the hybrid-face robot. (d) Hybrid-face robot's thirteen degrees of freedom: left and right eyebrow angle $B_{al}$ and $B_{ar}$; left and right eyebrow vertical height $B_{hl}$ and $B_{hr}$; left and right eyelid openness $L_l$ and $L_r$; eye pitch and yaw $E_p$ and $E_y$; pupil size $P$; mouth corner vertical height $M_h$; mouth width $M_w$; top lip openness $M_t$; and bottom lip openness $M_b$ for emotion depiction.

The hybrid-face robot consists of eyebrows, eyelids, eyeballs, and mouth, with a total of thirteen degrees of freedom (DoF) (Fig. 1(d)). These thirteen values characterise the facial expression at any given time. The hybrid-face was programmed in the OpenGL environment and rendered using Face3D.

We propose two types of affect space representations to generate emotions for the hybrid-face robot – categorical affect space and three-dimensional affect space, described below. We also added eye blinks, subtle twitching and constant motion of eyes to make the hybrid-face robot more dynamic, expressive, realistic and likeable. Further details of the hybrid-face robot affect space design summarised below are given in our previous work [44].

*1) Categorical Affect Space Emotion Representation*

The categorical affect space represents the robot's facial expression by a linear combination of its basis expressions. Our set of basis expressions, extended from Breazeal's work [45], consists of *happy, sad, angry, afraid, surprise, tired, stern,* and *disgust* ($B = \{\vec{b}_1, \vec{b}_2, ..., \vec{b}_n\}$), each of which is a vector containing thirteen values corresponding to thirteen degrees of freedom of a hybrid-face.

An expression $\vec{e}$, is created by a weighted linear combination of variances of different expressions from neutral expression ($\vec{b}_i - \vec{b}_N$) added to the neutral expression $\vec{b}_N$.

$$\vec{e} = \left(\sum_{i=1}^{n}(\vec{b}_i - \vec{b}_N)\vec{w}_i\right) + \vec{b}_N$$

Where, $n$ is a number of basis expressions, weight vector with a weight corresponding to each basis expression is $\vec{w} = [w_1, w_2, ..., w_n]$, $w_i \in [0,1]$. Fitzpatrick [46] and Bruce [47] have shown that such emotions can be used in sophisticated human-robot interactions.

*2) Three-Dimensional Affect Space Emotion Representation*

We developed a three-dimensional affect space for the hybrid-face robot inspired by Breazeal [45]. Unlike typical two-dimensional affect space with *arousal* and *valence* axes [48, 49], we use three dimensions with *arousal, valance,* and *stance* axes (Fig. 2), capable of capturing the vast majority of facial expressions [48, 50]. The three axes are characterised by six basis expressions $B = \{\vec{b}_{happy}, \vec{b}_{sad}, \vec{b}_{surprise}, \vec{b}_{tired}, \vec{b}_{angry}, \vec{b}_{afraid}\}$, with pairs of opposite expressions on each end of axis: valence axis with happy and sad, arousal axis with surprise and tired, and stance axis with angry and afraid. Each expression is a linear combination of three basis expressions, one on each of the axes.

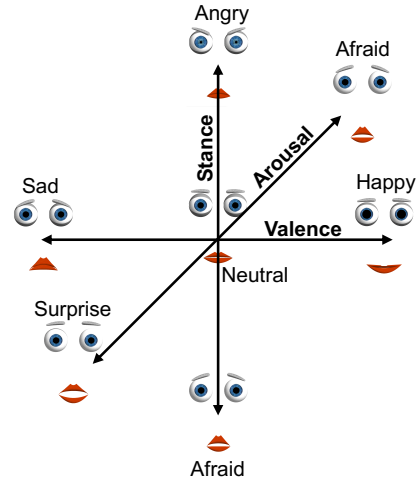

Fig. 2. Three-dimensional affect space represented by axes of arousal (high or low), valence (positive or negative) and stance (open or closed).

The representation of an expression $\vec{e}$ at a location $\vec{x} = [\alpha\ \beta\ \gamma]^T$, $\alpha, \beta, \gamma \in [-1, +1]$ in the 3D affect space is given by the linear combination of variances of basis expressions along three axes from neutral expression ($\vec{b}_i - \vec{b}_N$) as follows:

$$\vec{e} = \max(\alpha, 0)(\vec{b}_{happy} - \vec{b}_N) + \max(-\alpha, 0)(\vec{b}_{sad} - \vec{b}_N) +$$
$$\max(\beta, 0)(\vec{b}_{surprise} - \vec{b}_N) + \max(-\beta, 0)(\vec{b}_{tired} - \vec{b}_N) +$$
$$\max(\gamma, 0)(\vec{b}_{angry} - \vec{b}_N) + \max(-\gamma, 0)(\vec{b}_{afraid} - \vec{b}_N) + \vec{b}_N$$

where the maximum function *max* negates the contribution of a basis expression if $\vec{x}$ is closer to its opposite basis expression. Unlike Breazeal's work, we have placed angry and afraid expressions on the positive and negative end of the stance axis to simplify the affect space.

*B. Participants*

To assess conscious and subconscious responses to the hybrid-face robot, we performed tests with 19 healthy participants (22 - 58 years, 9 female and 10 male). This group of participants was selected from a wide age range and diverse professions to better resemble the population that social robots would have to engage with. All 19 participants participated in behavioural response experiments out of which 12 participants also took part in neurophysiological response experiments. Ethical approval for the study was obtained from the University of Bristol Ethics Committee and all participants gave informed written consent.

*C. Behavioural Response to Hybrid-Face Robot*

*1) Emotion recognition experiment*

A forced-choice expression recognition task [12, 51, 52] was conducted for qualitative assessment of recognition of the hybrid-face robot emotions. Participants were shown different expressions (happy, sad, angry, afraid, surprise, tired, stern and disgust) in a random order for 4 s each and were asked to select the best matching emotion from a provided list of above eight emotions. Several instances of each expression were repeated with different combinations of static expression, expression with realism features (blinks and twitches) and expression with animation i.e. transitioning from the neutral expression.

*2) Pupil dilation experiment*

This experiment investigated the relationship between pupil dilation and implied emotion from the hybrid-face to quantify empirical values of pupil dilation for each emotional expression. We aimed to investigate whether pupil dilation values improve emotion recognition rates if the robot face showed pupil dilation size that humans may innately associate with a 'universal standard' for each emotion.

Seated 1 m away from the robot, participants were not asked to recognise an emotion for the expression but provide their subjective opinion to three pupil dilation sizes; the minimum (when the pupil dilation began to suit the face), the target (when the pupil dilation matches the expression) and the maximum (when the pupil dilation begins to mismatch with the facial expression). To quantify participant's subjective expectations of pupil dilation to the eight robot facial expressions, each expression was presented initially with a minimal pupil dilation that increased over time to a maximum. The digital pupil dilation of the robot was increased gradually at a rate of 0.6 mm/s from a minimum 10 mm diameter to a maximum 40 mm diameter, on a white sclera of 85 mm diameter surrounding a blue iris with a 45 mm diameter.

*D. Neurophysiological response to Hybrid-Face Robot*

Drawing from physiological studies on the quantification of neural response to human emotion, EEG experiments were conducted to validate the subconscious response of humans to robot emotions. The purpose was to investigate if robotic emotion conveyance could evoke comparable neurological responses to human emotion conveyance such as face-specific N170 ERP. To validate our ERP experimental paradigm and provide a baseline with N170 response evoked by human faces, we conducted a pre-pilot with a subset of three participants (see Appendix I). Participants were shown pictures of human faces [53] with different standardised emotions according to FACS on a monitor and their EEG was recorded.

*1) Experimental Design*

Participants seated 1 m away from a monitor and hybrid-face robot (Fig. 3), were naïve to the research question and were asked to observe stimuli. Neurophysiological response to structured facial expressions and the effect of emotion presentation with context (hybrid-face robot) or without context (digital face on monitor) was studied. For the experiment with the face on a monitor, a fixation cross was presented for a random duration around 2 s followed by an emotion stimulus in random order for 1 s. A blank screen was displayed for 1 s between trials. The experiment with hybrid-face robot followed the same structure, only the stimulus was presented on the hybrid-face robot which provided additional configurational information. Twenty-five EEG trials were recorded for each emotion during robot and monitor conditions.

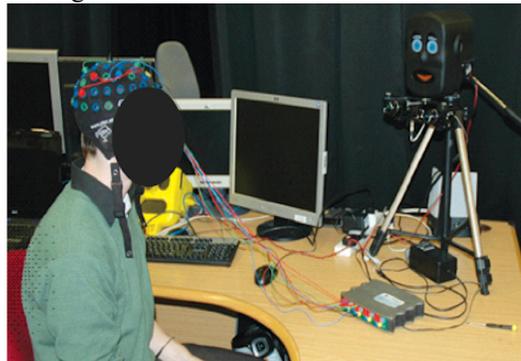

Fig. 3. EEG experiment setup with the participant seated in front of hybrid-face robot and monitor.

*2) EEG Recording*

EEG was recorded using a 16 channel 24bit g.tec USBAmp (g.tech medical engineering, Schiedlberg, Austria), sampled at 256 Hz with online bandpass-filtering between 0.10-30Hz. Electrodes were placed at Fp1, Fpz, Fp2, F3, Fz, F4 (frontal), C3, Cz, C4 (central), T7, T8 (temporal), P7, P3, Pz, P4, P8 (parietal) and Oz (occipital) locations according to 10-20 international system with reference at left ear lobe and ground at Fpz. Electrode impedance was kept below 20 kΩ.

*3) ERP Analysis*

EEG was filtered between 0.1-20 Hz using a zero-phase shift digital low pass filter. Artefacts were removed by rejecting trials with amplitude greater than ±70 µV on channels Fp1 and Fp2 and by visual inspection. The artefacts removed EEG was segmented into epochs -100 ms to 400 ms after the stimulus onset. The grand average ERP was extracted by averaging of the time locked trials across subjects and each stimulus. We characterised the latency of N170 ERP component by averaging percentiles from minimum amplitude within 130-190 ms post-stimulus time window.

*E. Results of Hybrid-Face Robot Experiments*

*1) Behavioural response to the hybrid-face robot*

Table. I shows the confusion matrix of average emotion recognition accuracies from forced-choice experiments under all conditions (static, realism, animation and their combination). Happy, sad and surprise expressions had the highest recognition accuracies. However, stern was confused with tired and angry, angry was confused with stern, afraid was confused with sad, and disgust was confused with surprise, possibly due to the similarity between the expression features. Poor identification of emotion disgust is well noted in the literature [24-26], however, we observed lowest accuracies for afraid and stern with the hybrid-face robot.

TABLE I. CONFUSION MATRIX OF OVERALL EMOTION RECOGNITION ACCURACIES (%) FOR HYBRID-FACE ROBOT

|          | Happy | Sad  | Angry | Afraid | Surprise | Tired | Stern | Disgust |
|----------|-------|------|-------|--------|----------|-------|-------|---------|
| Happy    | 84.8  | 1.3  | 2.7   | 1.3    | 4.6      | 4.0   | 1.3   | 0       |
| Sad      | 1.5   | 88.8 | 0.6   | 3.9    | 1.3      | 0.6   | 2.0   | 1.3     |
| Angry    | 3.3   | 2.7  | 68.5  | 3.3    | 0.6      | 0.6   | 21.0  | 0       |
| Afraid   | 2.0   | 28.2 | 2.6   | 50.0   | 9.9      | 3.3   | 2.0   | 2.0     |
| Surprise | 2.7   | 2.0  | 1.3   | 9.2    | 79.6     | 3.2   | 2.0   | 0       |
| Tired    | 0     | 3.2  | 6.6   | 4.0    | 2.0      | 69.0  | 14.5  | 0.7     |
| Stern    | 0     | 6.0  | 10.7  | 0.7    | 4.0      | 14.0  | 59.2  | 5.4     |
| Disgust  | 4.6   | 4.6  | 0     | 0      | 15.8     | 6.6   | 4.6   | 63.8    |

Table. II shows correct emotion recognition percentages under animation, realism and static conditions. Animation showing transition of emotion from neutral expression decreased recognition rates for angry, afraid, tired and disgusted, possibly due to unnatural speed of animations.

TABLE II. EMOTION RECOGNITION ACCURACIES (%) FOR HYBRID-FACE ROBOT IN STATIC, ANIMATION (TRANSITIONING FROM NEUTRAL) AND REALISM (BLINKS AND TWITCHES) CONDITIONS

|                      | Happy | Sad  | Angry | Afraid | Surprise | Tired | Stern | Disgust |
|----------------------|-------|------|-------|--------|----------|-------|-------|---------|
| Overall              | 84.8  | 88.2 | 68.5  | 50.0   | 79.6     | 69.0  | 59.2  | 63.8    |
| Static               | 89.5  | 89.5 | 76.3  | 50.0   | 84.2     | 76.3  | 51.2  | 65.8    |
| Animation            | 79.0  | 81.6 | 55.3  | 44.8   | 86.8     | 63.2  | 57.9  | 57.9    |
| Realism              | 86.9  | 89.5 | 71.0  | 55.3   | 84.2     | 71.1  | 65.8  | 68.4    |
| Animation & Realism  | 84.2  | 92.1 | 71.0  | 50.0   | 63.1     | 65.81 | 60.5  | 63.2    |

*2) Effect of pupil dilation on emotion recognition*

The results for pupil dilation experiment, recorded as the percentage of iris diameter, are presented in Fig. 4(a). The pupil dilation for a given expression was recorded for three interest points; minimum, target and maximum pupil dilation.

The pupil dilation for the happy expression is consistently greater than neutral expression across the three interest points. Surprise is the only other emotion to provide an equally large dilation percentage comparable with happy. Angry has consistently the smallest pupil size across all interest points. The ideal target size of pupil dilation for the neutral expression was consistent at 25% of the iris diameter.

Results indicating the difference between expected pupil dilation for emotional face diagrams as a percentage of total iris diameter relative to neutral, the relationships can be suggested; happy (+3 to +8%), stern (-2%), angry (-3 to -5%), afraid (-3%), sad (+4%) and disgust (+1 to +3%). The results for tired vary inconsistently across interest points. Results for the three interest points suggesting these relationships are presented as percentages of the total iris size to allow for comparison and across emotions (Fig. 4(b)).

The results across the three interest points for other emotions vary around neutral and between happy and angry expressions.

While small differences between grand-average results for each emotion can be due to with-in variation, an overall relationship between pupil dilation and the happy, angry and neutral face diagrams is demonstrated.

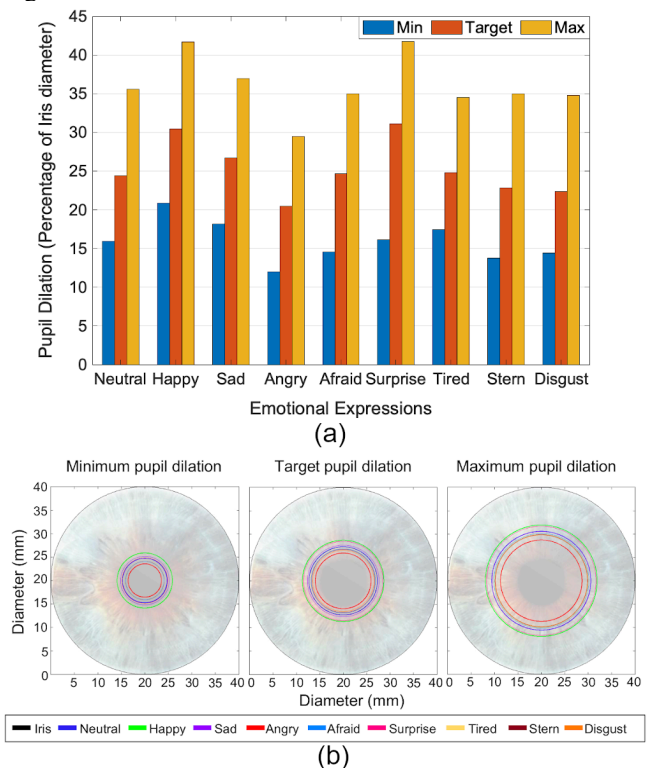

Fig. 4. Pupil dilation results. (a) Grand-average results for pupil dilation expectations to match emotion for three interest points: minimum, target and maximum presented as percentage of the total iris size. (b) Grand-average results for pupil dilation to match emotion for three interest points illustrated on the image of iris for comparison.

*3) Physiological ERP response to the hybrid-face robot*

A clear face-specific N170 ERP was observed in the parietal region for hybrid-face robot emotions. The grand-average N170 ERP amplitudes and waveforms are presented in Fig. 5 for each emotion at the P8, for both robot and screen sources. This ERP response was similar to N170 ERP response to pictures of human facial expressions as described in Appendix I.

Statistical significance of different expressions and effect of context presentation on ERP was assessed using Analysis of Variance (ANOVA). Facial expressions with strong positive or negative emotions evoked a larger ERP response than neutral. Angry emotion showed highest amplitude (Angry$_{MONITOR}$ ($F(1,259)=3.03$, $p=0.083$), Angry$_{ROBOT}$ ($F(1,259)=4.43$, $p=0.036$). The differences in the amplitude and latencies of N170 for different emotions within the same stimuli presentation source (monitor or robot) were not statistically significant. However, differences in amplitude of N170 between the two stimuli presentation sources were statistically significant ($F(1,17)=11.73$ $p<0.01$). The topography of EEG showing N170 in the two stimuli presentation sources for different emotions is shown in Appendix II. An average delay of 7 ms was also observed between two presentation sources.

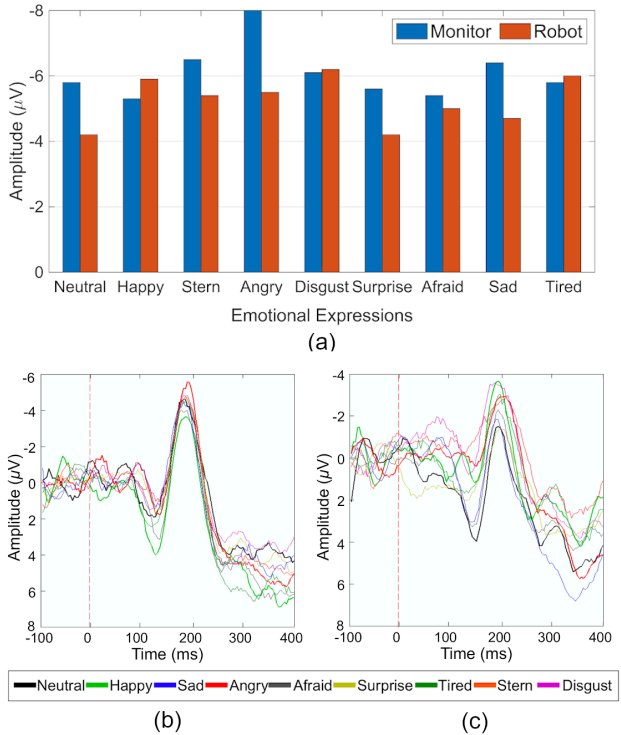

Fig. 5. N170 ERP response to hybrid-face robot expressions. (a) Comparison of N170 ERP amplitudes between stimulus sources (monitor and hybrid-face robot) for all emotions at P8. (b) Grand-averaged ERP waveforms for digital expressions presented on the monitor (without context) at P8. (c) Grand-averaged ERP waveforms for digital expressions presented on the hybrid-face robot (with context) at P8.

## III. Translation of Hybrid-Face System for Mass-Market Social Robot

The hybrid-face robot concept has been validated for its capacity to convey different emotions successfully as shown by the qualitative assessment with strong recognition rates by humans and by quantitative assessment of face-specific N170 ERP neurophysiological responses in the brain. These results provide a basis for full translation of the hybrid-face concept in the development of a commercial social robotic platform. The next subsections detail the development and testing of an application of a hybrid-face in a IoT social robot dubbed Miko 'My Companion', executed by Emotix Inc (Mumbai, India). Despite its simplicity, a full hybrid-face robot with all 3D facial features was not practical as a stand-alone device for mass-market. A further simplified version of the face and expressions was necessary without compromising recognition rates and human response to emotions, while improving likeability.

### A. Design of Miko Social Robot

Miko robot is an intelligent IoT based social robot designed for educational purposes (https://miko.ai/in). Miko draws upon the hybrid-face robot concept to improve human-robot affective interaction. The first-generation robot we have produced, Miko I, is simplified from the original hybrid-face robot and consists of eyes which can show different emotions to facilitate communication. The hybrid face of Miko I itself is placed on its head which also has ears and a curved surface, giving an illusion of depth (Fig. 6). Design choices such as the curved hybrid-face, colour, ears, and shape were chosen through qualitative experiments with customers and tools such as Quality Functional Deployment (QFD). Development of Miko I was guided by a user centric design approach involving feedback from over 300 young school students. Students were able to identify majority of the Miko expressions and always correctly identified whether the emotion associated to the displayed expression was positive or negative. Sound cues helped identification of emotions; however, a clear visual representation of expression was found to be more important. Communication with Miko I occurs through IoT connectivity via an app that allows users to talk and send various emotions using different emoticons. Miko I displays received emotions along with audio and light stimuli as well as small movements matching the emotions. The range of emotions for Miko I, drawing from the hybrid face, are shown in Fig. 6.

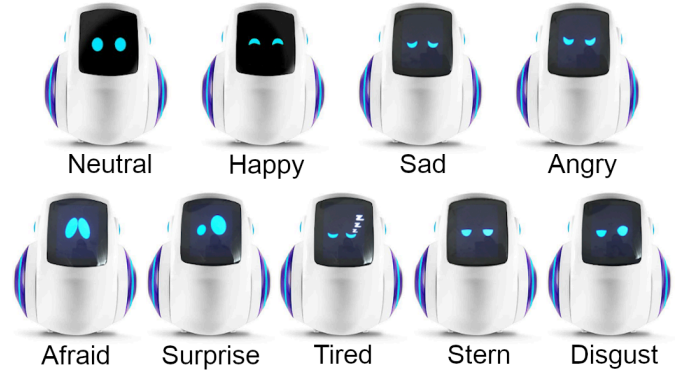

Fig. 6. Miko I social robot with a curved display resembling a 3D structure of a face showing different emotions.

In this study, we focus on the recognition and neurophysiological response of emotions displayed on the hybrid-face of Miko I to validate its capacity to evoke human emotive response. Having established a procedure for response validation, we execute this procedure with Miko I and compare its response to that of the full hybrid-face robot with 13 DoF.

The depiction of emotions of Miko I is simplified and do not show the gradation/transition depending on parameters as with the full hybrid-face robot. While more complex transitions and emotions may be incorporated in future releases, we wish to test if the simplified emotions in Miko I provide comparable conveyance and recognition to human-human interaction via behaviour and neurophysiological (N170 ERP) studies.

### B. Participants

To assess conscious and subconscious responses to Miko I robot similar to hybrid-face robot, we performed tests with two groups of healthy participants. A group of 15 participants (19 - 29 years, 5 female and 10 male) participated in the behaviour response experiments with Miko I and a second group of 10 participants (22 - 29 years, 1 female and 9 male) participated in the neurophysiological response experiments. Ethical approval for the study was obtained from Imperial College London Science, Engineering and Technology Research Ethics Committee and all participants gave informed written consent.

### C. Behavioural Response to Miko I Robot

We repeated the behavioural analysis with emotion recognition task on Miko to qualitatively assess recognition of different expressions shown by Miko. The experiment structure was the same as the forced-choice experiment conducted with the hybrid-face robot described in section II.C.1. To avoid response bias, subjects who had never interacted with robot Miko were selected. Participants were given a list of the same emotions and after each emotion shown by Miko, they were asked to select the best matching emotion. Emotions were sent to Miko manually from its companion mobile app and each emotion was displayed for 4 s with several repetitions in random order. In this experiment, the movement of Miko was constricted but the sound and light stimulus occurring along with the facial expressions were kept on, that might help in emotion recognition. The recognition rate of different Miko expressions were recorded.

### D. Neurophysiological response to Miko I Robot

The aim of this experiment was to study whether Miko I robot with simplified facial features shows face-specific N170 ERP neurophysiological response to different emotions similar to the full hybrid-face robot.

#### 1) Experimental design

Miko I was placed approximately 90 cm away from the participants on a desk. The experiment setup is similar to the one shown in Fig. 3. Neurophysiological responses to four emotions: angry, happy, sad and surprised were tested. We selected these four emotions because they are far apart on the three-dimensional affect space axes (see Fig. 2) and showed strong N170 ERPs with the hybrid-face robot. The emotions were sent to Miko I via its companion mobile app manually at the beginning of each EEG trial. During each EEG trial, Miko I displayed the emotion for 4 s followed by a break of 4 s. The order of the sent emotions was randomised to avoid anticipation of the next emotion. A camera co-registered with EEG recording was placed facing Miko I that recorded Miko's emotions which was later used to extract the exact time of onset of the stimulus (emotion) shown by Miko I. During a 4 s break period, Miko showed neutral expression and blinked regularly. The movement of Miko was restricted and the lights and sounds presented by the Miko during different emotions were switched off as they provide additional multimodal stimuli.

#### 2) EEG Recording

EEG was recorded using TMSi Porti amplifier and EEG cap with passive electrodes (TMSi, Oldenzaal, The Netherlands). 16 unipolar channels of EEG were recorded from the locations Fp1, Fp2, Fpz, F3, F4, Fz (frontal), C3, C4, Cz (central), T7, T8 (temporal), P3, P4, Pz, Poz (Parietal) and Oz (occipital) according to the 10-20 international system. Channel AFz was used as common ground. EEG was recorded at 2048 Hz and downsampled to 256 Hz during analysis. 60 EEG trials were recorded for each of the four emotions for each participant.

#### 3) ERP Analysis

EEG was filtered between 0.1-45 Hz using 4th order zero-phase shift band-pass filter to remove DC offset and high-frequency noise. Artefacts were removed using Independent Component Analysis from EEGLAB toolbox [54]. The independent components containing mostly ocular artefacts were identified manually and removed. The artefacts removed EEG was segmented into epochs -100 ms to 400 ms after the stimulus onset. The stimulus onset was extracted from the video of Miko emotion changes. The video was recorded at 26 fps and hence the stimulus resolution obtained was 38.46 ms. The mean of the 100 ms pre-stimulus was used as the baseline for normalisation. Any trial with an amplitude above ±70 µV was excluded from further analysis. Each trial was then filtered between 1-5 Hz to obtain the ERPs. The N170 ERP for each emotion was extracted through the grand average of time locked EEG trials across all subjects.

### E. Experimental Results of Emotive Response to Miko I

#### 1) Behavioural response to Miko I

Table. III shows the confusion matrix of Miko I emotion recognition accuracies by participants in percentage during forced-choice emotion recognition experiment. The expressions happy, sad, angry and tired showed the highest recognition rates. Other expressions showed lower recognition rates. Particularly, participants confused stern with disgusted more than half of the times and showed the lowest recognition rates. Other emotions that were generally confused were stern with tired, disgusted with surprised, and surprised with afraid. Interestingly, the recognition rate of emotion afraid in Miko I showed significant increase compared to the hybrid-face robot.

TABLE III. CONFUSION MATRIX OF EMOTION RECOGNITION ACCURACIES (%) FOR MIKO ROBOT

|  | Happy | Sad | Angry | Afraid | Surprise | Tired | Stern | Disgust |
|---|---|---|---|---|---|---|---|---|
| Happy | 93.3 | 0 | 0 | 0 | 0 | 0 | 0 | 6.7 |
| Sad | 0 | 93.3 | 0 | 0 | 6.7 | 0 | 0 | 0 |
| Angry | 6.7 | 0 | 93.3 | 0 | 0 | 0 | 0 | 0 |
| Afraid | 0 | 6.7 | 0 | 80.0 | 13.3 | 0 | 0 | 0 |
| Surprise | 0 | 0 | 0 | 6.7 | 73.3 | 0 | 0 | 20.0 |
| Tired | 0 | 0 | 0 | 0 | 0 | 86.7 | 13.3 | 0 |
| Stern | 0 | 0 | 6.7 | 6.7 | 0 | 6.7 | 40.0 | 40.0 |
| Disgust | 0 | 0 | 0 | 6.7 | 6.7 | 6.7 | 46.7 | 33.3 |

Overall the results of recognition were in agreement with the results of the hybrid-face robot. Recognition accuracies were higher for happy, angry and sad and lower for stern and disgust than the corresponding recognition accuracies for the hybrid-face robot. We hypothesize this small difference is a result of loss of mouth, which might add critical information for recognition of stern and disgust.

#### 2) Physiological ERP Response to Miko Robot

All participants showed distinct changes in neurophysiological markers in response to Miko I robot depicting different emotions. We observed ERP component N170, which is consistent with identification of faces in EEG in response to Miko I emotions. Grand-average ERP responses to different emotions in channels Pz, Poz and Oz are shown in Fig. 6. Strong ERPs were observed in parietal regions as expected. This shows that the Miko I robot with simplified facial features also successfully evoked responses that are typically obtained by observing fully articulated face with all the features such as mouth and eyebrows, which is the direct response documented to human facial expressions. Emotion surprised showed strongest N170 in Poz and Oz, whereas happy had highest N170 in Pz. ERPs of different emotions between 150 – 250 ms were not significantly different as determined by ANOVA, consistent with the hybrid-face robot. Thus, responses to Miko I were similar to those observed for the hybrid-face robot and simplification of facial features did not decorate recognition or emotional response.

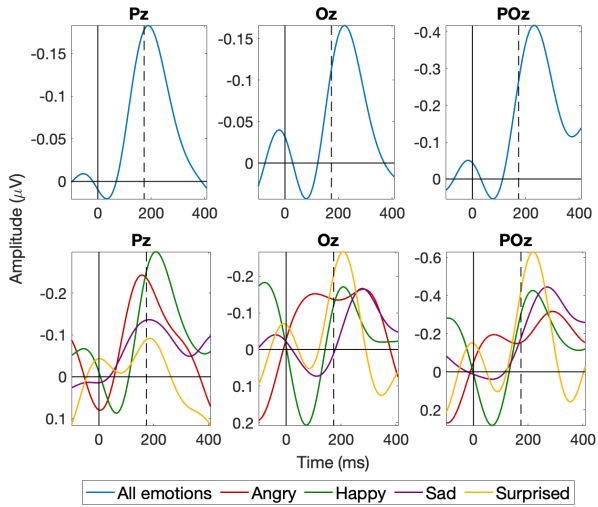

Fig. 7. N170 ERP responses to Miko I expressions. (a) Grand-average N170 ERP waveform for all emotions together at Pz, Oz and POz. (b) Grand-average N170 ERP waveform for individual emotions at Pz, Oz and POz.

## IV. Discussion

### A. Summary of findings

In this study, we designed, validated emotional response, simplified, and commercially translated a hybrid-face robotic system. We investigated and quantified human physiological response to digital facial expressions from two versions of the robot: a hybrid-face robot research platform and Miko I, its simplified version commercial release.

Human-robot interaction and cooperation requires trust. Trust is typically achieved in human interactions using emotions and similar approach could enhance human-robot engagement by enabling robots to convey emotions via facial expressions. An optimal balance must be achieved between realistic appearance and iconic appearance of robot. The state space for developing affective robot face is huge, hence based on pragmatic approach, we have made assumptions to simplify this space to create a hybrid-face robot that gives illusion of depth without compromising the simplicity of design and implementation. Since developing a fully actuated face was not practical, we simplified it by developing a hybrid-face with similar features. This hybrid-face was still not practical for commercial applications and hence we simplified it even further to develop Miko I robot, which still captured the core of human-robot interaction as contrasted to established human-human responses (face-specific N170 ERP).

We assume that there are different levels of abstraction for developing a facial robot in increasing order as follows: human face, fully actuated face, hybrid face, simplified hybrid face and digital avatar. Intuitively, people respond better to a face that shows depth. Hence, we designed hybrid-face robots that gave perception of depth with faceplate without actuation on the face. By studying participants' conscious response and then evaluating their subconscious response (through ERP), we found that not actuated depth-based robots give comparable human-like affective responses as seen from similar face-specific N170 ERP, which is a well-established response to human faces [39]. Our experimental design also allowed participants to move their head and explore three-dimensionality of the robot which can be paralleled to real human interaction. Including depth in the design of hybrid-face robot helped in building the context, which is important to enhance human-robot interactions. Though we found the depth in representation of the robot's face is useful pragmatically to create better affective experience, further work will be needed to quantify its effect.

### B. Hybrid-face robot validation

First, we examined the utility of hybrid-face robot with 3D printed faceplate and digital display as a platform for human-robot engagement. As an attempt to verify the functionality of the hybrid-face robot, and to gain insights into mathematical representations of affective potential, emotion recognition experiments were carried out. Participants were able to identify different emotions with high accuracies. The animation and realism features could be fine-tuned in the future work to increase expression recognition. The hybrid-face design is very flexible and adaptable to incorporate new expressions.

We demonstrated the ability of the digital facial expressions to effectively convey emotion to a human observer by recording event-related potentials in EEG to determine the perception of digital emotion. We found significant difference in ERP due to the context presentation using monitor and the hybrid-face robot, thus depth shows differences in physiological response to face. We ensured that participants were naive to the research questions to avoid bias in face processing known to occur due to context and manipulation by an emotionally laden task [55].

A distinctive N170 component was reliably identified in the grand-average response from all participants, for all hybrid-face robot emotions. The digital facial expressions were able to modulate the ERP response despite having low information bandwidth. Our results are consistent with previous studies of the human response to conveyed emotion which state both pleasant and unpleasant expressions evoked a larger N170 ERP response than neutral expression and this change in activity is located on the right parietal brain area (Appendix II Fig. 9). These results are in line with other trials conducted with human images (including our ERP validation in Appendix I Fig. 8), though our results for the robot's facial expressions are modulated with a lower amplitude and increased latency, congruent with a similar study by Dubal [56], who found that robot expressions are encoded as early as human faces but evoke a later and muted response. This indicates that several neuron clusters associated with internal features and head detection maybe engaged requiring additional time to acquire configural information. Positive emotions were evoked earlier than negative, which follows the current literature. This work confirms the trend by extending physiological responses to human expressions produced by more than forty-four muscles on the face to very simplified robot expressions.

An average delay of ~7 ms occurred in physiological responses between the two sources of computer monitor and the hybrid-face robot. This might be caused due to the increase in configural information or a delay as attention is refocused to the hybrid face. While a fixation cross was presented on the monitor, subjective comments from participants noted that focusing attention was easier and more natural for the hybrid face with head and ears, which helped setting the configuration parameters.

These results show considerable promise because participants were able to identify most of the expressions and responded positively to hybrid-face robot interaction. Participants quickly accepted digital facial expressions causing their attention to evolve beyond robot's physical characteristics. Thus, we propose a novel method to quantify human-robot engagement using empirical N170 ERP measure.

Pupil dilation is an often overlooked dimension of non-verbal communication that can influence perceived emotion. We added pupil dilation to the hybrid-face robot in the attempt to improve conveyance of emotions by increasing the degrees of freedom without adding actuation based on our pragmatic assumptions. Inclusion of pupil dilation feature is inconclusive currently and it is difficult to determine whether it helps with the confidence of emotion recognition and requires further investigation. Participants noted that they were not aware of how dramatic the effect of pupil size was on the overall demeanour and interpretation of robot's expressions. Varying the size of the iris may impact the expected dilation of the pupils for different emotional expressions. The current eye colour of the digital face provides a definite contrast between the light blue iris and black pupil, that even with increased pupil dilation, gives the appearance of a cold stare for some facial expressions. The results from this experiment provide average estimations of pupil dilation to robot's facial expression that will help improve emotion recognition rates in further experiments.

*C. Simplification of Hybrid-Face Robot for mass market*

Acceptability of engaging with the hybrid-face robot by human participants and high recognition rates of different emotions presented by the hybrid-face robot was promising. However, the hybrid-face approach was still not viable for incorporating in a commercial product which required further simplified face. Thus, the hybrid-face robot influenced the development of a commercial social robot Miko I capable of affective conversations by creating further abstractions of emotion representation. Here, we studied Miko's ability to convey different emotions and human response to those emotions and compared those with the results of the hybrid-face robot by repeating the same set of experiments. Even though Miko I had simplified facial features with just two eyes, the behavioural and physiological experiments showed comparable results to a full hybrid-face robot. Thus, Miko I was able to convey emotions successfully using static, singular expressions with eyes only, simplifying affect space representation. Miko I has integrated IoT capabilities to enable affective engagement with children for educational purposes. The conversational ability of Miko I which is not investigated in this study may benefit greatly from using the appropriate expressions to enrich the affective information content in the conversation and enhance engagement with humans. Similar to hybrid-face, participants reported that the curvature and ears on Miko I representing facial structure providing contextual information helped in associating with the humanoid form and in recognising expressions.

The physiological study of human responses to Miko emotions also showed a distinct face-specific N170 ERP component during all the emotions. This validates the findings of simplified hybrid-face principle and its translation to Miko I. The amplitude of N170 was smaller because our EEG itself had an overall smaller amplitude. The latency of N170 could not be estimated accurately because the stimulus was extracted from a video with a resolution of 38.46 ms. Strongest ERP was obtained for emotion surprised which was represented by bigger eye size than other emotions, however, participants showed some uncertainty in recognition of this emotion. Thus, experiments with Miko showed a successful practical application of hybrid-face in affective social robot with IoT and demonstrated evoked response to human-robot affective interaction on a physiological level.

Based on these findings, we argue that simplified robotic platforms fusing static mechanical design and digital encoding can evoke conscious and subconscious emotive response in human beings comparable to human-human interaction.

## V. Conclusions

We have presented a complete project life cycle, from concept, to design, implementation, testing, validation, and commercial translation of hybrid-face robotic system capable of evoking human-like affective response in users. A first generation affective hybrid-face robot is developed to help researchers design intelligent systems capable of ensuring mutual trust, safety, and effective cooperation with humans. We successfully applied this hybrid-face to develop a commercial IoT social robot, Miko I, by providing it with the ability to integrate affective information in its interactions. The qualitative and quantitative assessment of both the hybrid-face robot and its application to Miko I demonstrated that human participants were able to recognise the emotions conveyed by the robots with high accuracy and also showed physiological face-specific N170 ERP response in their EEG which is typically obtained with observing human faces. This validated the effectiveness of emotion conveyance by social robots.

In summary, new contributions of this investigation include:
- Establishing that 'hybrid face' robots are comparable to human faces for emotional state conveyance
- Derivation of a canonical set of facial degrees of freedom, including pupil dilation, for deployment of affective robotics in real-world environments with IoT connectivity
- Introduction of two models - categorical and affect space - for representing robotic expressions
- Introduction of the use of empirical recording of EEG physiological response as a tool to quantitatively assess human response to robotic emotions; demonstration that EEG mapping of subconscious response to human-robot visual affective interaction correlates to that of human-human visual affective interaction
- Presented a useful approach for practical commercial translation of affective IoT human-robot interface systems leading to a new mass market robotic product

Emotix has, since the time of these experiments, released a new robot with greater autonomy (Miko II) drawing on these findings. Future work will involve implementing Miko II and other forms of the hybrid-face robot to accelerate learning in children and as a support tool for the elderly in isolation.

## APPENDIX

### I. COMPARISON OF N170 ERP RESPONSE TO HYBRID-ROBOT FACE AND PICTURES OF HUMAN FACES

Fig. 8 shows comparison of N170 ERP response to hybrid-face robot and pictures of human facial expressions taken from Japanese Female Facial Expression database [53] with FACS expressions in three participants. This was conducted with a small subset of participants to validate our ERP paradigm and observe whether human faces show the expected face-specific N170 response. A strong N170 ERP is obtained for human faces as well as for the hybrid-face robot.

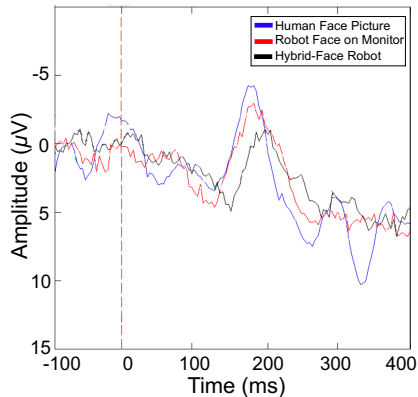

Fig. 8. Comparison of grand-averaged N170 ERP response to digital face presented on monitor and hybrid-face robot, and pictures of a human face.

### II. EEG TOPOGRAPHY MAP OF DIFFERENT STIMULI

Fig. 9 outlines multiple views of EEG topography showing spatial location of N170 ERP response to neutral, happy, sad and angry expressions for computer monitor and robot stimuli sources. The digital expressions presented on monitor evoked larger N170 ERP as compared to expressions on robot.

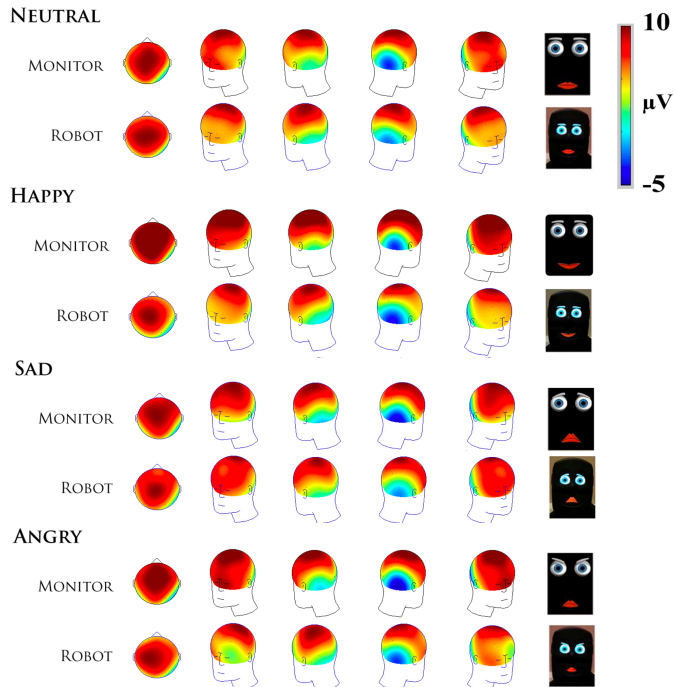

Fig 9. EEG topography of evoked negative N170 ERP response to expressions neutral, happy, sad and angry presented on monitor and hybrid-face robot.

## ACKNOWLEDGEMENTS

The authors would like to thank all participants involved in this study, as well as Emotix Inc for their innovative drive in introducing the Miko platform and support of translational research.